\documentclass[sn-mathphys,Numbered]{sn-jnl}

\usepackage{graphicx}%
\usepackage{caption}
\usepackage{subcaption}
\usepackage{comment}
\usepackage{multirow}%
\usepackage{amsmath,amssymb,amsfonts}%
\usepackage{amsthm}%
\usepackage{mathrsfs}%
\usepackage[title]{appendix}%
\usepackage{xcolor}%
\usepackage{textcomp}%
\usepackage{manyfoot}%
\usepackage{booktabs}%
\usepackage{algorithm}%
\usepackage{algorithmicx}%
\usepackage{algpseudocode}%
\usepackage{listings}%
\usepackage{booktabs} 
\usepackage{caption} 

\theoremstyle{thmstyleone}%
\theoremstyle{thmstyletwo}%

\theoremstyle{thmstylethree}%
\newtheorem{definition}{Definition}%

\begin{document}



\title[Article Title]{Redefining Event Types and Group Evolution in Temporal Data}


\author[1,2]{\fnm{Andrea} \sur{Failla}}

\author[3]{\fnm{Rémy} \sur{Cazabet}}

\author[2]{\fnm{Giulio} \sur{Rossetti}}

\author*[2]{\fnm{Salvatore} \sur{Citraro}}\email{salvatore.citraro@isti.cnr.it}

\affil[1]{\orgdiv{Department of Computer Science}, \orgname{University of Pisa},  \city{Pisa}, \country{Italy}}

\affil*[2]{\orgdiv{Institute of Information Science and Technologies “A. Faedo” (ISTI)}, \orgname{National Research Council (CNR)} \city{Pisa}, \country{Italy}}

\affil[3]{\orgname{Univ Lyon, UCBL, CNRS, INSA Lyon, LIRIS, UMR5205, F-69622}, \city{Villeurbanne}, \country{France}}


\abstract{
Groups --- such as clusters of points or communities of nodes --- are fundamental when addressing various data mining tasks.
In temporal data, the predominant approach for characterizing group evolution has been through the identification of ``events".
However, the events usually described in the literature, e.g., shrinks/growths, splits/merges, are often arbitrarily defined, creating a gap between such theoretical/predefined types and real-data group observations.
Moving beyond existing taxonomies, we think of events as ``archetypes" characterized by a unique combination of quantitative dimensions that we call ``facets".
Group dynamics are defined by their position within the facet space, where archetypal events occupy extremities.
Thus, rather than enforcing strict event types, our approach can allow for hybrid descriptions of dynamics involving group proximity to multiple archetypes.
We apply our framework to evolving groups from several face-to-face interaction datasets, showing it enables richer, more reliable characterization of group dynamics with respect to state-of-the-art methods, especially when the groups are subject to complex relationships.
Our approach also offers intuitive solutions to common tasks related to dynamic group analysis, such as choosing an appropriate aggregation scale, quantifying partition stability, and evaluating event quality.
}

\keywords{group evolution, temporal clustering, community detection, clustering evaluation}

\maketitle
\section{Introduction}
\label{sec:intro}

Unsupervised learning, such as clustering and community detection, involves identifying collections of elements that share some form of similarities.
Clustering methods identify groups of observations or entities, based on their proximity across multi-dimensional features \cite{macqueen1967some}.
Community detection aims to describe the mesoscale dimension of a complex network, grouping nodes that share similar structural behaviour \cite{fortunato2016community}.
Such groups are often referred to as clusters in data mining-related fields, and communities in complex network analysis. In the remainder of this article, we will use the generic term of ``group".
Groups are fundamental when addressing a wide variety of data mining and network science-related questions, such as segmenting customers to improve recommender systems or identifying echo chambers in social media to de-polarize online discussions.

Often, real data are stored as streams or sequences of multi-dimensional points \cite{zubarouglu2021data} or links \cite{rossetti2018community} such that the formation and evolution of groups can be tracked quantitatively over time.
The challenge of identifying evolving groups has emerged as a distinct subfield in many areas; see spatiotemporal data clustering \cite{kisilevich2010spatio, ansari2020spatiotemporal}, data stream clustering \cite{zubarouglu2021data}, and dynamic community detection \cite{rossetti2018community}.

A key concept specific to dynamic group evolution is the notion of \emph{event} \cite{palla2007quantifying,lughofer2012dynamic, brodka2013ged,rossetti2018community}.
Intuitively, an event is a temporal occurrence involving changes that can be measured and analyzed for a group or a set of groups.
A change can be the growth or the shrinking of a group, or the merging of two groups into a larger one. 
These events have been defined in the literature \textit{a priori} \cite{greene2010tracking, asur2009event, brodka2009performance, brodka2013ged, palla2007quantifying}, based on what one wishes to extract from the data, and not from the reality of group evolution observed in data.
As a consequence, in most datasets we are confronted with a gap between theoretical events such as ``merge" or ``growth", and what one actually observes. 
Most group evolution seems indeed more complex, frequently being a combination of those artificial categories.



In this work, rather than using those strict event definitions, we consider them as ``archetypes", i.e., typical examples of a category conveying its most salient features \cite{rosch1975cognitive}, while real events can exhibit features from multiple of these archetypes.
To tackle this more complex definition, we propose a quantitative definition of event archetypes as a unique combination of three constitutive dimensions called \textit{facets}.
Each event is thus defined by a position in this 3-dimensional space, in which usual events occupy an extremity.
Following \cite{bovet2022flow}, we consider \emph{backward} and \emph{forward} perspectives to study the temporal evolution of a target group.

The rest of the paper is organized as follows.
Section \ref{sec:related} sums up the essential literature about temporal clustering across different domains and the need to build taxonomies to describe the life cycles of groups.
Section \ref{sec:methods} describes our framework for characterizing the temporal evolution of target groups as weighted approximations of archetypal events.
Section \ref{sec:exp} introduces an experimental setting to test our methodology in real-world data.
Finally, Section \ref{sec:disc_concl} concludes the work by discussing its potentiality and limits.

\section{Related Work}
\label{sec:related}

\begin{figure}
     \centering
     \begin{subfigure}[b]{0.3\textwidth}
         \centering
         \includegraphics[width=\textwidth]{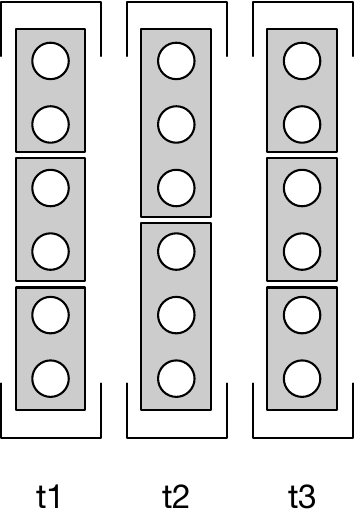}
         \caption{Sequence of partitions}
         \label{fig:1a}
     \end{subfigure}
     \hfill
     \begin{subfigure}[b]{0.3\textwidth}
     \includegraphics[width=\textwidth]{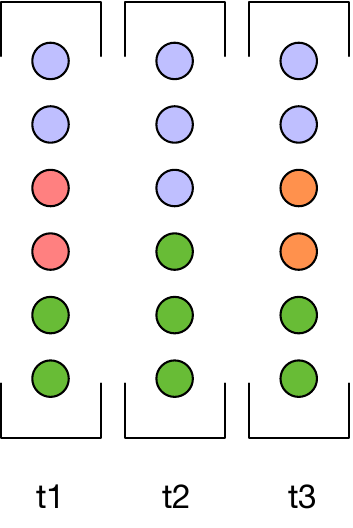}
         \caption{(Entity/time clusters)}
         \label{fig:1b}
     \end{subfigure}
     \hfill
     \begin{subfigure}[b]{0.3\textwidth}        \includegraphics[width=\textwidth]{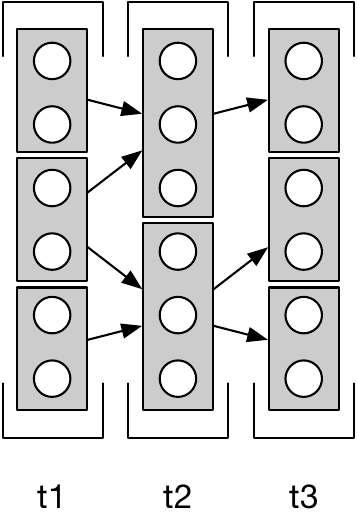}
         \caption{Event graph}
         \label{fig:1c}
     \end{subfigure}
        \caption{Three representations for dynamic groups found in the literature. a) Does not describe the relation between the groups; b) Assigns labels (here represented by colors) to each entity/time, yielding a longitudinal group; c) Describe how groups in a timestep are related to those in the next. Note that none of these representations explicit the events occurring on the network.}
        \label{fig:representations}
\end{figure}

Data stream clustering, temporal clustering, and dynamic community detection are active research topics, and many algorithms have been proposed to identify time-varying groups \cite{rossetti2018community,kisilevich2010spatio,ansari2020spatiotemporal,zubarouglu2021data}. However, most of these works do not address the question of group events. 
Indeed, dynamic grouping methods often yield their groupings in one of two forms, as illustrated in Fig. \ref{fig:representations}: (i) The method might only focus on the groups found at each step, for instance ensuring the stability of these groups in time. 
In that case, we know what are the groups at $t$ and at $t+1$, but there is no information in the relation between groups in $t$ and in $t+1$ (Fig. \ref{fig:1a}); or (ii) each group is a set of (entity/time) pairs, i.e., a group exists over multiple timesteps, potentially allowing entities to join and leave the group along time (Fig. \ref{fig:1b}). However, none of these representations explicit what the events undergone by evolving groups are.
In this work, we will consider starting from a sequence of temporally ordered observations, i.e., snapshots, and sets of partitions on these snapshots, obtained from an existing method --- or possibly, from a ground truth --- as illustrated in Fig. \ref{fig:1a}. The task consists in characterizing the nature of the relation between groups at time $t$ and at time $t+1$, in the form of events.

The earliest attempts to define events on groups can be found in Kalnis et al. \cite{kalnis2005discovering} and Hopcroft et al. \cite{hopcroft2004tracking}, with the objective to identify \textsc{Continue} events, i.e., to define that community $c_1$ at time $t_1$ should be considered as the \textit{same community} as $c_2$ at time $t_2$, a process also known as \textit{matching} groups.
Kalnis et al. \cite{kalnis2005discovering} proposed to match groups if (i) they are adjacent, and (ii) their Jaccard coefficient is above a threshold $\tau$. 
Given $c_t$ and $c_{t+1}$, namely two groups observed over temporally adjacent snapshots, if $\frac{|c_t \: \cap \: c_{t+1}|}{|c_t \: \cup \: c_{t+1}|} \ge \tau$, the two groups are the same cluster.
$\tau$ is a parameter considering the ``integrity" of a cluster, and it indicates the minimum overlap threshold required for two clusters to be considered the same.
Similarly, to identify a \textsc{Continue} event in dynamic networks, Hopcroft et al. \cite{hopcroft2004tracking} defined a match function as follows: $match(c_t,c_{t+1})=Min(\frac{|c_t \: \cap \: c_{t+1}|}{|c_t|}, \frac{|c_t \: \cap \: c_{t+1}|}{|c_{t+1}|})$; then, if $match(c_t,c_{t+1}) \ge \tau$, the two clusters are matched. However, without particular constraints on the value of $\tau$, there is no guarantee that a group is matched to a single community in the following step. The process thus naturally creates an \textit{event graph}~\cite{rossetti2018community}, as illustrated in Fig. \ref{fig:1c}. Greene et al.~\cite{greene2010tracking} were the first to formalize this event graph and to propose to use it to define typical events. A group having an out-degree of two or more is labeled as undergoing a \textsc{Split} event. An in-degree of two or more is a \textsc{Merge}. In/Out degrees of zero respectively lead to \textsc{Birth} and \textsc{Death} events. Finally, the authors mention that \textsc{Contraction}, \textsc{Expansion}, and \textsc{Continue} all correspond to the same graph setting, but can be distinguished by adding size thresholds. It is worth noting though that despite its elegance and apparent simplicity, this approach creates complex situations: a group $(t_1,c_1)$ can be matched to two groups $(t_2,c_2)$,$(t_2,c_3)$, but one of these groups might itself have an in-degree of more than one, thus being a merge of $(t_1,c_1)$ and $(t_1,c_4)$. These situations are not discussed in the article of Greene et al~\cite{greene2010tracking}.

This merge-split ambiguity can be solved by adding some additional constraints, such as in Asur et al. \cite{asur2009event}, in which an event is categorized as a \textsc{merge} if the following three conditions hold: (i) $\frac{|(c^i_t \: \cup \: c^j_t) \: \cap \: c_{t+1}|}{Max(|(c^i_t \: \cup \: c^j_t)|,|c_{t+1}|)} > \tau$, 
(ii) $|c^i_t \cap  c_{t+1}|> \frac{|c^i_t|}{2}$, 
and (iii)  $|c^j_t \cap  c_{t+1}|> \frac{|c^j_t|}{2}$
where $c^i_t$ and $c^j_t$ are two distinct groups within the same partition at time $t$. 

Brodka et al. \cite{brodka2013ged} aimed to define as many events as possible by proposing a methodology for detecting group evolution, namely the Group Evolution Discovery (GED) framework.
They introduce a measure to quantify the \textit{inclusion} of one group into another based on both group sizes and a centrality measure called Social Position (SP) \cite{brodka2009performance}, namely a function calculating how much a node is important in a group based on the importance of its neighbors.
The group events are defined on the basis of a decision tree that assigns an event type to a pair of groups given some thresholds on sizes and inclusion values.
Similarly to GED, Gliwa et al. \cite{gliwa2012identification} introduced an algorithm for Stable Group Changes Identification (SGCI), where complex events such as merge, split, and split+merge (a split of the original group and the joining of many groups into successor groups) are associated to groups that have been found stable in neighboring time steps. The stability depends on a match function defined as $Max(\frac{|c_t \: \cap \: c_{t+1}|}{|c_t|}, \frac{|c_t \: \cap \: c_{t+1}|}{|c_{t+1}|})$.
\\ \ \\
To complete the overview of the analysis of complex event types in data mining- and network science-related tasks, it is worth mentioning works moving beyond merely identifying group events, but also employing them to predict the future evolution of a system.
Typically, this subject has been modeled as a machine learning task, commonly in the form of a classification problem, where a set of features is extracted from the events and used to predict an event type \cite{saganowski2015predicting}.
The events defined for building the training sets can be obtained from the methods previously described.
In \cite{saganowski2015predicting}, events are described through the previously mentioned GED \cite{brodka2013ged} and SGCI \cite{gliwa2012identification} algorithms.
Group features used for prediction are size, density, and the sum/average/min/max of aggregated group members' features, such as node degree and other centralities measures.
Other works exploit forecasting methods for time series, predicting how the features extracted from the events will change in the following time period \cite{ilhan2015predicting}.
The latest directions of research are starting to focus on the behavior and prediction of individual entities.
For instance, \cite{tsoukanara2021should} proposed to predict, using node embedding methods, beyond event types, whether a specific node stays in the same group, moves to another group, or drops out of the dataset.

\begin{figure}
     \centering
     \begin{subfigure}[b]{0.6\textwidth}
         \centering
         \includegraphics[width=\textwidth]{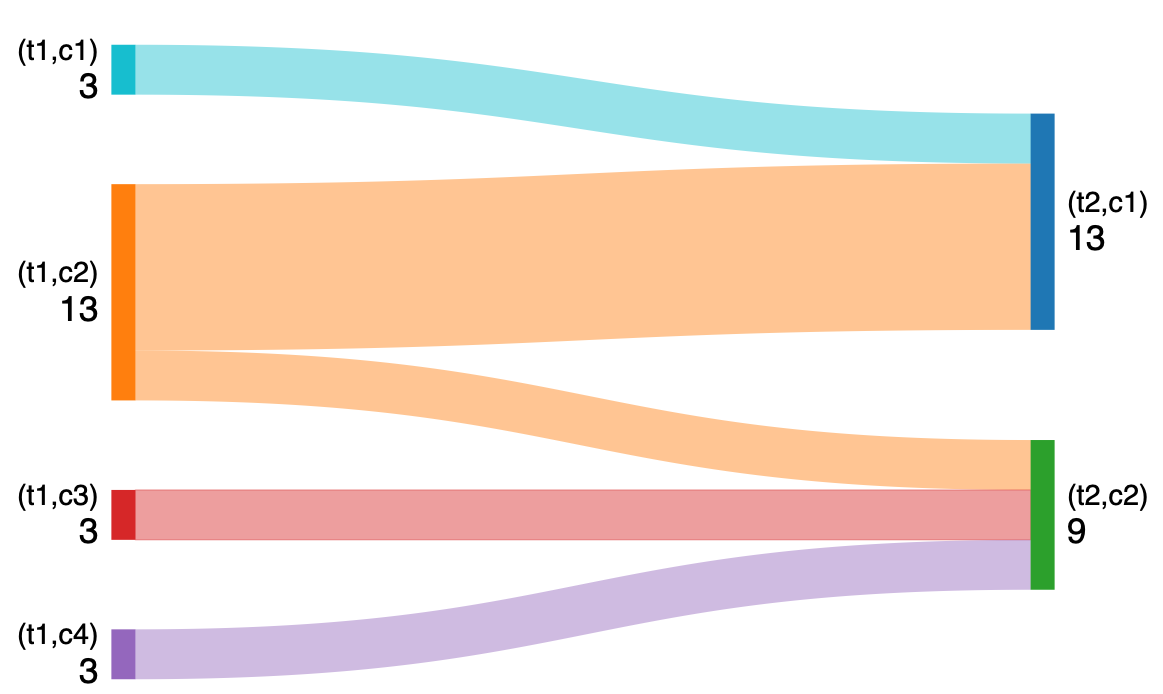}
         \caption{}
         \label{fig:complex1}
     \end{subfigure}

     \centering
     \begin{subfigure}[b]{0.15\textwidth}
     \includegraphics[width=\textwidth]{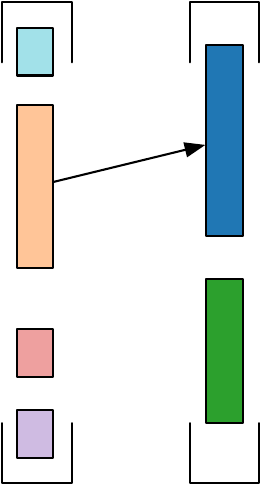}
         \caption{}
         \label{fig:complex2}
     \end{subfigure}
     \hspace{3em}
    \begin{subfigure}[b]{0.15\textwidth}
     \includegraphics[width=\textwidth]{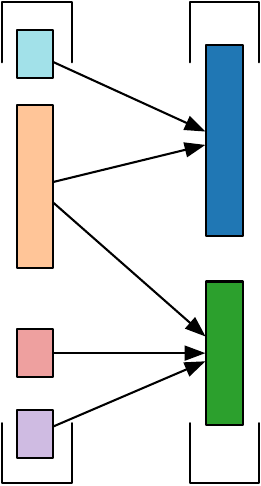}
         \caption{}
         \label{fig:complex3}
     \end{subfigure}
     \hspace{3em}
     \begin{subfigure}[b]{0.15\textwidth}
     \includegraphics[width=\textwidth]{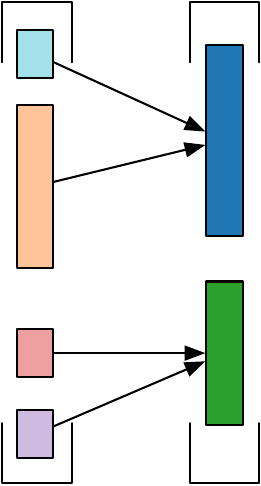}
         \caption{}
         \label{fig:complex4}
     \end{subfigure}
        \caption{A realistic group evolution scenario (a), with corresponding event graph with (b)intersection of union, high threshold, (c) intersection over union, low threshold, (d)intersection over min size}
     
        \label{fig:complex}
\end{figure}

\subsection{Limits of existing methods}
Existing event detection methods suffer from some limitations, which might explain their limited use in practice.

The first limit is the sensibility to the fixed threshold $\tau$. Choosing a large threshold will cause many groups to die and be born at each step, while a low one will create complex events involving multiple groups from $t$ and from $t+1$ together, as illustrated in Fig.\ref{fig:complex}. Choosing this parameter thus strongly affects the results. 

Another limit is the discrepancy between having a list of well-defined events ---Continue, Birth, Death, Merge, Split, Expansion, Contraction--- and the more complex reality of real situations, as illustrated in Fig.\ref{fig:complex1}. We can observe that from the point of view of $(t1,c2)$, the event can be interpreted as either a division or a continue, with a minor number of entities leaving the group. 
From the point of view of $(t2,c1)$, we can again interpret what is happening as a merge or as a continue, depending on the threshold fixing what quantity of entities is considered a negligible fraction. An even more ambiguous case is the relation between $(t1,c2)$ and $(t2,c2)$ since a minority fraction of entities leaves the first, but they represent a large fraction of the newly formed one. We can formalize the difficulties encountered by those previous methods by their definition of an event, which involves a set of groups from two adjacent timesteps, without a notion of direction. 
Instead, we argue that a split, for instance, only makes sense when following the direction of time, while a merge is characterized by considering the opposite direction. 
Thus, in the case of Fig. \ref{fig:complex3}, we must extract a split from the point of view of (t1,c2), in the direction of time, and a merge from the point of view of (t2,c1), in the reverse direction.

Finally, a last limit of those approaches comes from their disregarding of the distinction between entities moving between groups, and those joining or leaving the network altogether. As an example, a death event might actually correspond to multiple situations: (i) all the entities composing the group might have left the studied system, (ii) entities might have split up and joined other groups. Conversely, a newborn group might be formed from entities joining the system, or from individual entities coming from multiple sources. Intuitively, these situations are different, but existing events do not distinguish them.

\section{Methods}
\label{sec:methods}

\begin{table}[tp]
    \centering
    \begin{tabular}{ c c }
    \hline
     $U$ & universe set \\ 
     $U_t$ &  elements observed at time t \\  
     $\mathcal{S}$ &  temporally-ordered set of partitions \\  
     $\mathcal{S}_t$ &  partition at time t \\  
     $X$ & target set \\
     $\mathcal{R}$ & sets the target evolves into/from \\
     $R$ & a child/parent set of the target \\
     $\mathcal{U}$ & Unicity facet \\
     $\mathcal{I}$ & Identity facet \\
     $\mathcal{O}$ & Outflow facet \\
     $\mathcal{M}$ & Metadata facet \\ 
     $\mathcal{T}$ & Event Typicality Index \\ 
    \hline
    \end{tabular}

    \caption{Notation used in the paper}
    \label{tab:notation}
\end{table}

This section describes our framework for characterizing event types and group evolution in temporal data.

We decompose the method description into three sections: (i) A description of the forward/backward event perspective, (ii) The definition of event description scores called \textit{Event facet scores}, (iii) The definition of event names and archetypes from combinations of facet scores.

\subsection{Forward and Backward events}
Our method starts from the postulate that events are defined either as \textit{forward} or \textit{backward} \cite{bovet2022flow}. 
The former corresponds to events seen from the perspective of a group at $t$, relative to groups at $t+1$.
Conversely, the latter corresponds to the point of view of groups from $t+1$ relative to those at $t$. 
This distinction is already present in the event-graph formalism: a merge event is defined as having an in-degree $k^{in}>1$, while a split corresponds to an out-degree $k^{out}>1$ (see Fig. \ref{fig:1c}). 
As discussed in section \ref{bidirectional}, events as they are defined in the literature (e.g., \cite{palla2007quantifying,Cazabet2018,rossetti2018community}) are archetypes composed of a combination of forward events and backward events. For instance, if a group has $k^{out}=1$ (i.e., forward perspective), we must consider $k^{in}$ (i.e., backward perspective) of the target group to know what this particular event is (continue or merge, in the event graph formalism).

\subsubsection{Definition}

Let $U = \cup_{t \in T} U_t$ be the universe set where each $U_t \subseteq U$ for $t\in T$ identifies the subset of elements of $U$ observed at time $t$. 
Let $\mathcal{S} = \{S_0, S_1, \dots, S_{|T|}\}$ be a temporally ordered set of elements where each $S_t = [S_t^0, S_t^1, \dots, S_{t}^m]$ identifies a partition of $U_t$.
Given a \textbf{target} set $X \in S_t$, let $\mathcal{R} = \{R^0,\dots, R^i\}$ be a \textbf{reference} subset of either $S_{t-1}$ or $S_{t+1}$ such that $X \cap R \neq \emptyset, \forall R \in \mathcal{R}$.
\\ \ \\
Following \cite{bovet2022flow}, the evolution of $X$ can be quantified by adopting either of two perspectives.
Under the \textit{backward perspective}, we look at the sets in $t-1$ that contribute to $X$'s formation; 
thus, imposing $\mathcal{R} \subseteq S_{t-1}$, we say that $X$ \textit{evolves from} $\mathcal{R}$. 
Conversely, under the \textit{forward perspective}, we look at the sets in $t+1$ that contain current $X$ members; thus, imposing $\mathcal{R} \subseteq S_{t+1}$, we say that $X$ \textit{evolves into} $\mathcal{R}$. 


\begin{figure}
     \centering
     \begin{subfigure}[b]{0.35\textwidth}
         \centering
         \includegraphics[width=\textwidth]{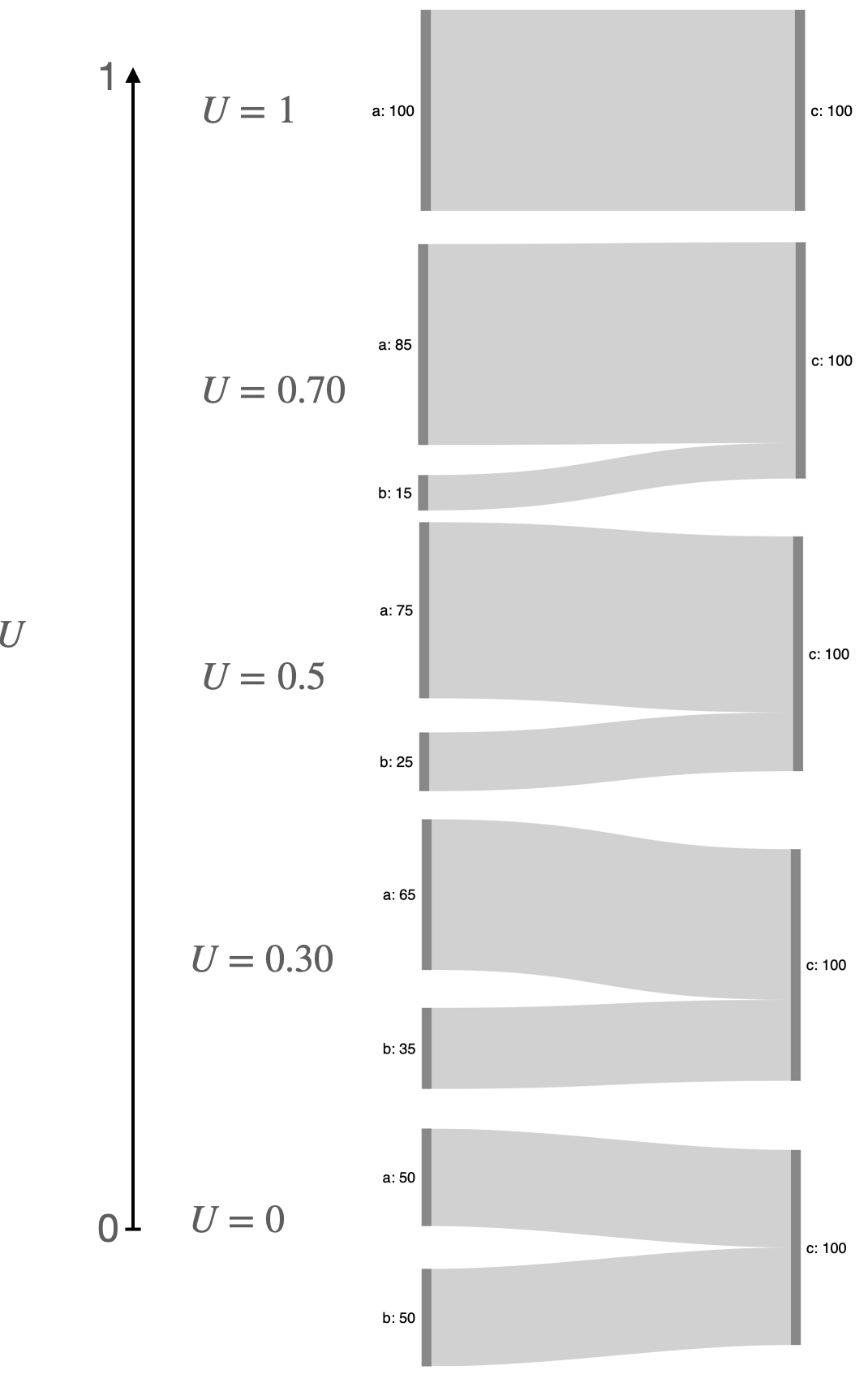}
         \caption{Unicity Facet $U$, for the right group, backward}
         \label{fig:Honly}
     \end{subfigure}
     \hspace{3em}
     \begin{subfigure}[b]{0.33\textwidth}
     \includegraphics[width=\textwidth]{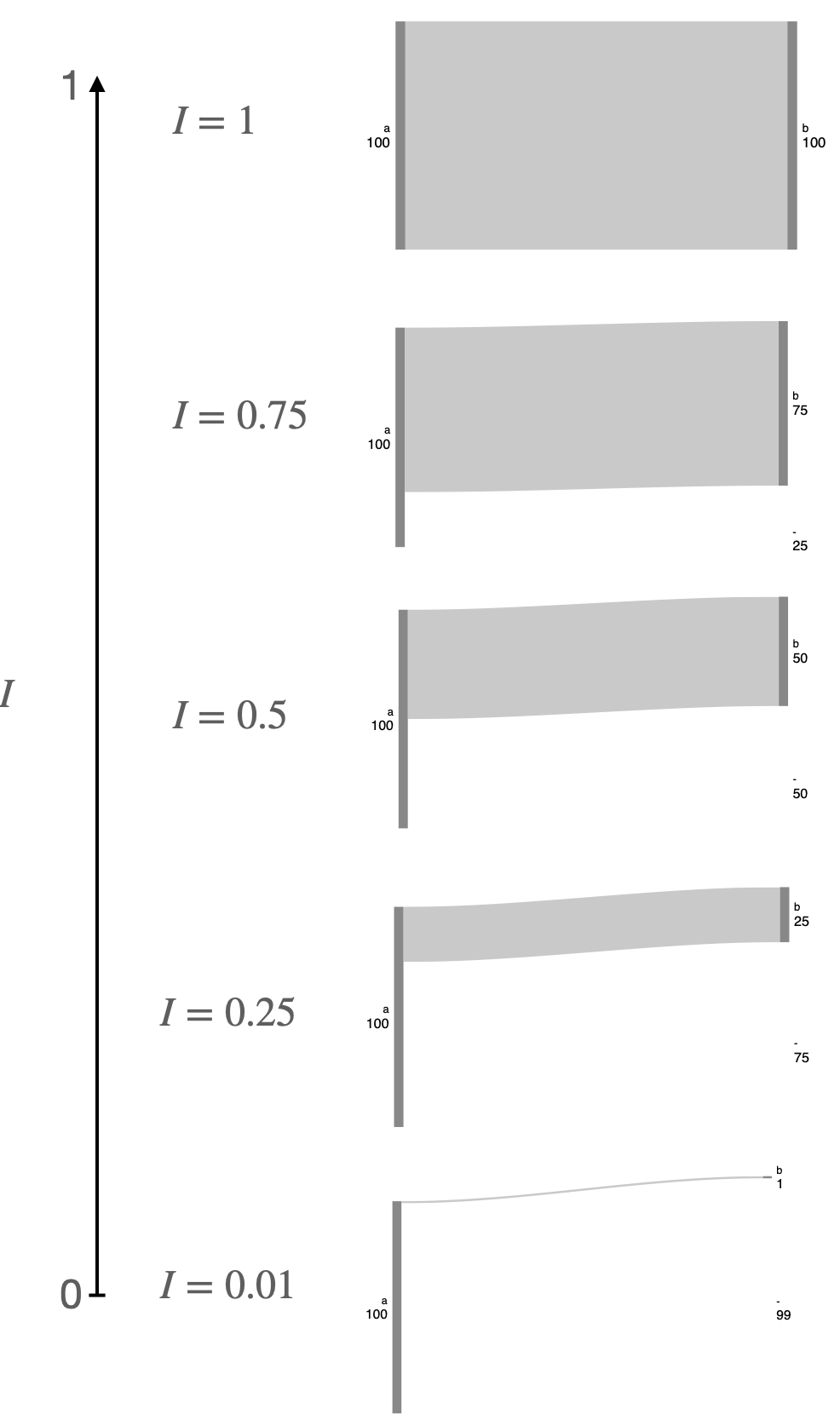}
         \caption{Identity Facet $I$, for the right group, backward}
         \label{fig:Wonly}
     \end{subfigure}

        \caption{Representation of the continuous nature of facets $U$ and $I$}
        \label{fig:facets}
\end{figure}

\subsection{Event facet scores}
In the literature, events are defined as mutually exclusive categories. In the event-graph representation, the in-degree of a group allows one to distinguish between a merge ($k^{in}\leq 2$), a death ($k^{in}=0$), and a continue or its variant ($k^{in}=1$). Instead, our approach first describes quantitatively the nature of an event using scores called \textit{event facets}. 

\subsubsection{Unicity Facet}
The first facet is called the Unicity Facet. In the forward perspective, it measures if the components tend to stay together or are disseminated in multiple destination groups. Conversely, in the backward perspective, it measures if all the entities come from a single source or multiple ones. The unicity facet can be understood as a continuous transition between $k^{in/out}=1$ and $k^{in/out}\geq2$ in the event graph formalism. Fig. \ref{fig:Honly} illustrates this facet in the backward perspective. It is computed as the difference between the fraction of elements in the two largest reference sets.

\vspace{2em}

\begin{definition}[Unicity Facet] 
\label{def:flowent}
Let $X$ be the target set, and $\mathcal{R} = [R_1,\dots, R_{|\mathcal{R}|}]$ be the reference set, ordered in decreasing order of intersection size with $X$, i.e., $\forall_{i> j},|R_i \cap X|\geq |R_j \cap X|$.

The Unicity Facet is defined as:
    \begin{equation}
    \label{eq:flowent}
        \mathcal{U} =     
        \begin{cases}
            \frac{|R_1\cap X|-|R_2\cap X|}{\bigcup_{R \in \mathcal{R}} R} & \text{if } |\mathcal{R}| \geq 2\\
            1  & \text{otherwise}
    \end{cases}
    \end{equation}
\end{definition}

This equation shares similarities with diversity indices such as the Gini's diversity index, and with dominance scores such as the Berger-Parker dominance score (later, BP), however, it also differs in important points. We can list the properties desired from $\mathcal{U}$ as follows:
\begin{enumerate}
    \item $\mathcal{U}=1$ if there is a single reference, as expected in a continue event.
    \item $\mathcal{U}=0$ if there are two or more references sharing equally the elements of $X$, as expected in an archetypal merge or split.
    \item Independently of $|\mathcal{R}|$, $\lim_{|R_1 \cap X|/|X| \to 1} \mathcal{U}=1$
    \item  Independently of $|\mathcal{R}|$, with $|R_1|=|R_2|$, $\lim_{|(R_1\cup R_2) \cap X|/|X| \to 1} \mathcal{U}=0$
\end{enumerate}
Property 1. is true for Gini and BP. It is true for $\mathcal{U}$ by considering that $|R_2|=0$ if $|\mathcal{R}|=1$. Property 2 is true for Gini but not for BP. Property 3 is true for BP but not for Gini, because it depends on $|\mathcal{R}|$. Property 4 is false both for Gini and BP. 

When no set contributes to $X$, i.e., all elements $x \in X$ are observed for the first time, it can be thought of as if all elements came from a single set existing before our observation period. We thus assign $\mathcal{U} = 1$ in that case.

\subsubsection{Identity Facet}

An aspect of events that is not directly tackled by existing literature is the question of the preservation of groups' identity. 
The term is taken as a parallel with social groups, such as for instance a political party. 
The group itself can be considered to have an identity, and if too many individuals leave the group, that identity might be lost. 
The Identity facet measures how much of the identity is transferred by/to the target group. 
Taking as an example the backward perspective, let us consider a group $g$, which receives all of its nodes from a group $g_{from}$. 
The Identity facet measures if the elements of $g$ represent a large fraction of the nodes of $g_{from}$, or only a small fraction of them. 
The facet thus ranges from $0$ to $1$, $1$ if all the nodes of $g_{from}$ joined $g$--- thus, the nodes are considered to come with the identity of their group of origin--- and tends towards 0 as the nodes joining $g$ represent a smaller fraction of $g_{from}$. Fig. \ref{fig:Wonly} illustrates this facet in the backward perspective.

\vspace{2em}

\begin{definition}[Identity Facet]
\label{def:contr}
    \begin{equation}
    \label{eq:contr}
         \mathcal{I} = \begin{cases}
         \frac{1}{|\bigcup_{R \in \mathcal{R}} R|} \sum_{R \in \mathcal{R}} |R \cap X | \frac{|R \cap X|}{|R|} & \text{if } |\mathcal{R}| > 0\\
         0  & \text{otherwise}
         \end{cases}
    \end{equation}
\end{definition}

\noindent As an illustrative example, assume that a single set of 10 elements, $R$, provides elements to $X$ with two alternative scenarios: a) it provides a single element, and b) it provides 9 out of 10 elements.
In the former scenario, the contribution $\mathcal{I}$ will approach 0; in the latter, it will approach 1 (reaching those extreme values only when none or all elements of $\mathcal{R}$ are present in $X$).

Note that $\mathcal{I}=0$ if there is no reference set: in that case, the group identity is completely lost (forward) or completely new (backward).

\subsubsection{Outflow Facet}

The outflow facet measures the fraction of the group elements that (i) just joined the system in the backward perspective, or (ii) left the system, in the forward perspective. 
The Outflow facet can be understood as a continuous transition between $k^{in/out}=0$ and $k^{in/out}=1$ in the event graph formalism, i.e., a birth (resp., death) and a continue. 
The facet thus ranges from $0$ --- all elements of the group were already present (resp., remain) in the system --- to $1$ --- all elements of the group are new (resp., are not present in the next timestep).

\vspace{2em}

\begin{definition}[Outflow Facet]
    \label{def:diff}
    \begin{equation}
    \label{eq:diff}
        \mathcal{O} = \frac{|X - \bigcup_{R \in \mathcal{R}}R|}{|X|}
    \end{equation}
\end{definition}


\subsubsection{Metadata facet}
Lastly, a fourth dimension can occur whenever entities or groups themselves are associated with some metadata.


We introduce the possibility of labeling the elements in a set with a categorical attribute $A$ such that $a(e) \in A$ identifies the categorical attribute value of an element $e$.
We assume the attribute value assigned to an element stays the same across time.
We use a measure of diversity \cite{morales2021measuring}, the Shannon entropy diversity index, to quantify changes in group mixing with respect to the attribute value.

\vspace{2em}

\begin{definition}[Attribute Entropy Change]
    Let the Attribute Entropy of $X$'s elements be:
    \begin{equation}
    \label{eq:attent}
    \mathcal{H}_{att}(X) =     
            -\sum_{x \in X}\frac{p(a(x))\log_2p(a(x))}{\log_2|\{a(x) \forall x \in X\}|} 
    \end{equation}
    The attribute entropy change is defined as the difference between the Attribute Entropy of the current set X and the mean of the Attribute Entropies of the reference sets. Formally:
    \begin{equation}
    \label{eq:attentch}
    \mathcal{M}=\Delta \mathcal{H}_{att} = \mathcal{H}_{att}(X) - \frac{1}{|\mathcal{R}|}\sum_{R \in \mathcal{R}}\mathcal{H}_{att}(R)
    \end{equation}

\label{def:attrent}
\end{definition}

\begin{figure}
     \centering
     \begin{subfigure}[b]{0.8\textwidth}
         \centering
         \includegraphics[width=\textwidth]{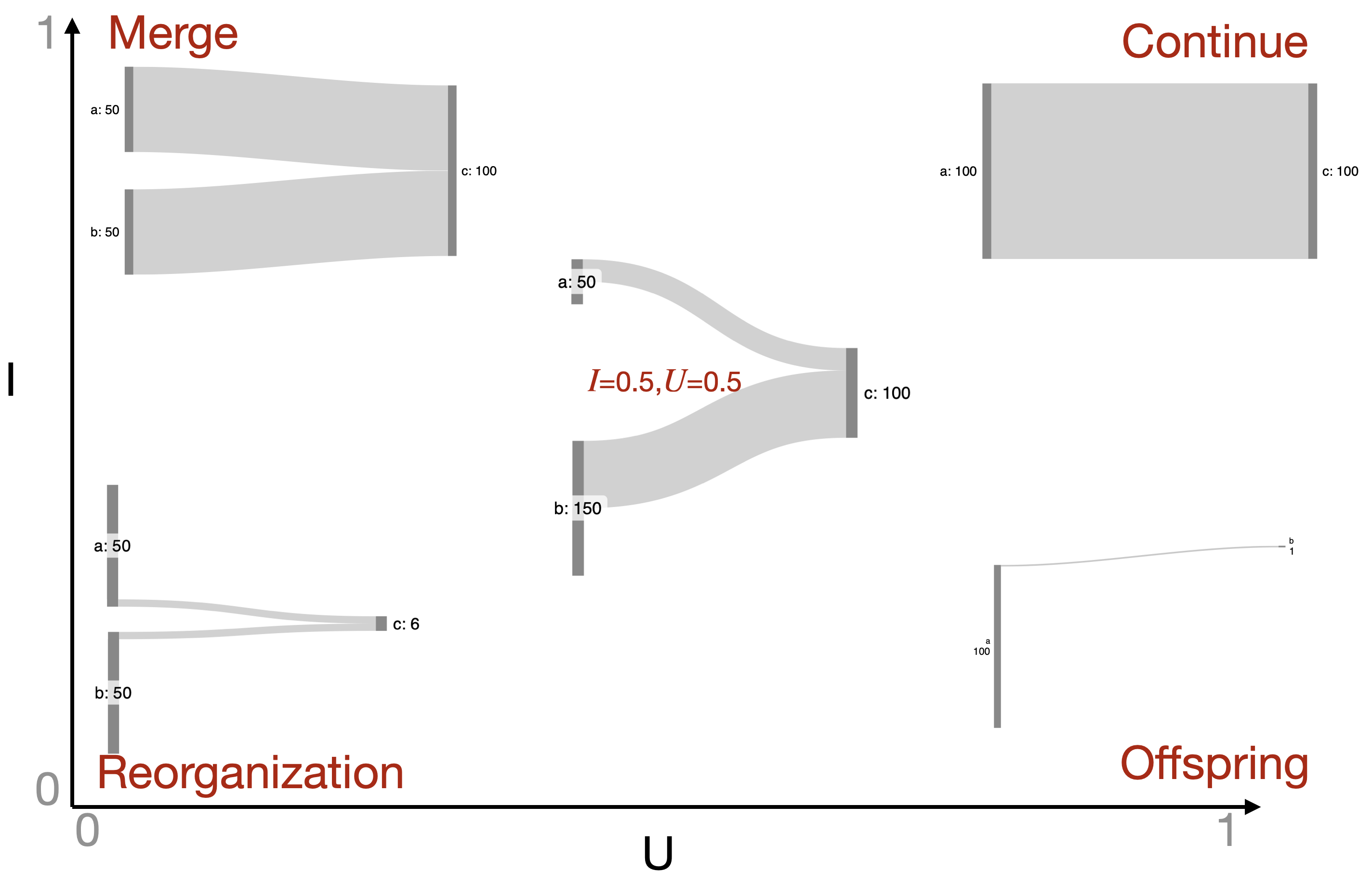}
     \end{subfigure}
        \caption{Archetype events according to values of facets $\mathcal{U}$ and $\mathcal{I}$, for $\mathcal{O}=0$. In the middle, an example of an event not clearly affiliated with an archetype according to these facets.}
        \label{fig:eventSpace}
\end{figure}

\subsection{Event weights and Archetypal events}



The facets introduced above describe intuitive quantities that can be used to characterize a group with respect to its evolutionary history, either past or future.
Going further, the evolutionary processes that outline a group's life cycle can be seen as a series of archetypal \textit{events} that the cluster undergoes. 
Here, we introduce the possibility of quantifying how much a group and its immediate predecessors/successors approximate some archetypal transformations, by combining facet scores into Event weights. Fig. \ref{fig:eventSpace} illustrates the relation between two of these facets and event archetypes, some of them common in the literature (Merge, Continue), some others being new (Reorganization, Offspring), but emerging naturally from our definitions.

We acknowledge that real-world evolutionary events are rarely found in their ``pure'' or ``archetypal'' form, often manifesting as complex and hybrid processes, often mired in messiness.
As such, it is relevant to characterize these processes as composite, measuring the extent to which they approximate (one or more) ``pure'' events.


Under the backward perspective, we can do so with the following \textit{event weights}.
\\ 
\begin{definition}[Backward Event Weights]
    \label{def:bweights}
    Let $X$ be the target set and $\mathcal{R}$ be the reference set such that $X$ evolves from $\mathcal{R}$.
    Backward event weights quantify the extent to which $X$'s evolution from $\mathcal{R}$ approximates one of the following transformations:

        \begin{align*}
        \textsc{Birth} = && \mathcal{U}\cdot(1-\mathcal{I}) \cdot \mathcal{O} \\
        \textsc{Accumulation} = && (1-\mathcal{U})\cdot(1-\mathcal{I}) \cdot \mathcal{O} \\
        \textsc{Continue} = && \mathcal{U}\cdot\mathcal{I} \cdot (1 - \mathcal{O}) \\
        \textsc{Merge} = && (1-\mathcal{U})\cdot\mathcal{I} \cdot (1 - \mathcal{O}) \\
        \textsc{Offspring} = && \mathcal{U}\cdot(1-\mathcal{I}) \cdot (1 - \mathcal{O}) \\
        \textsc{Reorganization} = && (1-\mathcal{U})\cdot(1-\mathcal{I}) \cdot (1 - \mathcal{O}) \\
        \textsc{Growth} = && \mathcal{U}\cdot\mathcal{I} \cdot \mathcal{O} \\
        \textsc{Expansion} = && (1-\mathcal{U})\cdot\mathcal{I} \cdot \mathcal{O} \\
        \end{align*}
\end{definition}

\textsc{Birth} events are characterized by a high number of joining elements that compose a set $X$, thus, a pure, archetypal \textsc{Birth} is found when the outflow $\mathcal{O}$ is maximized.
Theoretically, looking at the past to identify a birth is irrelevant since the appearance of new elements is unrelated to the incoming flow.
In real-world events, however, some fluctuating elements in $R$ can also join such newborn sets.
The Unicity facet $\mathcal{U}$ of the few elements joining a newborn set lets us further distinguish between a pure \textsc{Birth} and an \textsc{Accumulation}, i.e., a birth from subsets -- minimizing $\mathcal{U}$.
\\ \ \\
When the Identity facet is maximized in the absence of new elements, we characterize a \textsc{Continue}, if Unicity $\mathcal{U}$ is maximized, or a \textsc{Merge}, if $\mathcal{U}$ is minimized.
\textsc{Continue} events identify elements from a single set that are found together (i.e., in the same set) in the next timestamp.
\textsc{Merge} events identify the case where two or more sets of similar size join to form a single set in the next timestamp. 
\\ \ \\
When the Identity facet is low, and in the absence of new elements, we can witness an \textsc{Offspring} or a \textsc{Reorganization}, depending on the Unicity.
A pure, archetypal \textsc{Offspring} is observed when $\mathcal{U}$ is maximized, meaning that a small portion of a single origin set is found in the target set.
A \textsc{Reorganization} occurs when $\mathcal{U}$ is minimized, meaning the target set comprises small portions of several contributing sets.
\\ \ \\
When the past Unicity is maximized in the presence of many new nodes ($\mathcal{O} $ large), we can witness a \textsc{Growth}.
A pure, archetypal \textsc{Growth} is a single set that expands over the next timestamp, thus it is found by maximizing all three facets: Unicity (Single source), Identity (all the source), Outflow (numerous new nodes).
If $\mathcal{U}$ is minimized, we call the archetype an \textsc{Expansion}, i.e., similar to a \textsc{Growth} but from several contributing subsets. 
It should be noted that when $\mathcal{O}=1$, then by definition $\mathcal{U}=1$ and $\mathcal{I}=0$, i.e., we are in a \textsc{Birth} situation. This means that archetypes \textsc{Accumulation}, \textsc{Growth}, and \textsc{Expansion} can never reach a value of 1, but can only get close to this maximal value when a majority of nodes are new, but some come from existing groups.
\\ \ \\
\noindent Similarly to Definition \ref{def:bweights}, other events can be described by adopting the \textit{forward perspective}.
\\ \ \\
\begin{definition}[Forward Event Weights]
    \label{def:fweights}
    Let $X$ be the target set and $\mathcal{R}$ be the reference set such that $X$ evolves into $\mathcal{R}$.
    Forward event weights quantify the extent to which $X$'s evolution into $\mathcal{R}$ approximates one of the following transformations:

        \begin{align*}
         \textsc{Death} = && \mathcal{U}\cdot(1-\mathcal{I}) \cdot \mathcal{O} \\
        \textsc{Dispersion} = && (1-\mathcal{U})\cdot(1-\mathcal{I}) \cdot \mathcal{O} \\
        \textsc{Continue} = && \mathcal{U}\cdot\mathcal{I} \cdot (1 - \mathcal{O}) \\
        \textsc{Split} = && (1-\mathcal{U})\cdot\mathcal{I} \cdot (1 - \mathcal{O}) \\
        \textsc{Ancestor} = && \mathcal{U}\cdot(1-\mathcal{I}) \cdot (1 - \mathcal{O}) \\
        \textsc{Disassemble} = && (1-\mathcal{U})\cdot(1-\mathcal{I}) \cdot (1 - \mathcal{O}) \\
        \textsc{Shrink} = && \mathcal{U}\cdot\mathcal{I} \cdot \mathcal{O} \\
        \textsc{Reduction} = && (1-\mathcal{U})\cdot\mathcal{I} \cdot \mathcal{O} 
        \end{align*}
\end{definition}
\noindent The main difference with respect to the events in Definition \ref{def:bweights} relates to the meaning of $\mathcal{O}$. 
From the backward perspective, we compare the target set with the partition in the previous timestamp, so the elements quantified by the Outflow facet are ``new'', i.e., they are not present in the previous timestamp.
Contrarily, from the forward perspective, we compare the target set with the partition in the next timestamp, so the elements quantified by the Outflow Facet are ``dead'', i.e., they are not present in the next timestamp.
Thus the equation for \textsc{Death} is the same as for \textsc{Birth}.
A similar situation is found with \textsc{Accumulation} and \textsc{Dispersion}, \textsc{Merge} and \textsc{Split}, \textsc{Offspring} and \textsc{Ancestor}, \textsc{Reorganization} and \textsc{Disassemble}, \textsc{Growth} and \textsc{Shrink}, \textsc{Expansion} and \textsc{Reduction}.
The \textsc{Continue} event, instead, is undirected, meaning that it is measured in the same way regardless of the temporal direction.

\subsection{Event Typicality}
\label{typicality}
Finally, one might be interested in studying events that are closer to their archetype with respect to others, for instance, to separate them from more complex events. 
To distinguish between less and more pure transformations, we introduce the Event Typicality Index, defined as follows:
\\ \ \\
\begin{definition}
    Let $E_X$ be the set of Backward or Forward Event Weights computed for set X. 
    The Event Typicality Index $\mathcal{T}$ is computed as the maximum value among the event weights. 
    Formally:
    \begin{equation}
        \mathcal{T} = \max(E_X)
    \end{equation}
\end{definition}

Events corresponding perfectly to archetypes thus have $\mathcal{T}=1$.

\subsection{Bidirectional Events}
\label{bidirectional}

Events as they are usually defined in the literature, such as \cite{rossetti2018community}, are bidirectional in the sense that they are defined without a sense of direction. 
The exceptions are Birth and Death, which match our definitions, since they involve groups in only one timestep.
For the others, we can describe these bidirectional events using our formalism as a combination of forward and backward events, as follows: 
\begin{itemize}
    \item Literature continue corresponds to a forward \textsc{Continue} followed by a backward \textsc{Continue}.
    \item Literature merge corresponds to forward \textsc{Ancestor} events followed by a backward \textsc{Merge}. Note that if only two groups of equal size are involved, then the forward event will be between an \textsc{Ancestor} and a \textsc{Continue}
    \item Literature split corresponds to a forward \textsc{Split} event followed by an \textsc{Offspring} event. 
    As for merge, with only two groups, the backward events will be between an \textsc{Offspring} and a \textsc{Continue}
    \item Literature growth corresponds to a forward \textsc{Continue} followed by a backward \textsc{Growth} (or \textsc{Continue} if the faction of new nodes is small)
    \item Literature shrink corresponds to a forward \textsc{Shrink} (or \textsc{Continue} if the faction of quitting nodes is small) followed by a \textsc{Continue} event.
\end{itemize}

\section{Experiments}
\label{sec:exp}
In this section, we apply the proposed framework to analyze the evolution of groups from real-world data.

\subsection{Datasets}
We leverage the datasets from the SocioPatterns project\footnote{\url{www.sociopatterns.org}}, more precisely the Hospital~\cite{vanhems2013estimating}, Primary School~\cite{stehle2011high}, and High-School~\cite{mastrandrea2015contact} datasets. All of them correspond to face-to-face interactions collected via RFID sensors over 4, 2, and 7 days, respectively. 
They are frequently used in the context of dynamic network analysis, as they allow the consideration of multiple aggregation scales, have a reasonable size to be studied in detail, and undergo dynamics over the studied periods.

As a pre-processing, we first aggregate the data into a series of static networks, using a chosen time scale (hour, day, etc.). We then apply a community detection algorithm, namely the Louvain modularity maximization approach \cite{blondel2008fast}. This process yields sequences of node partitions, i.e., sets of sets, that constitute the input to our framework.

\begin{figure}
     \centering
     
    \begin{subfigure}[b]{0.31\textwidth}
     \includegraphics[width=\textwidth]{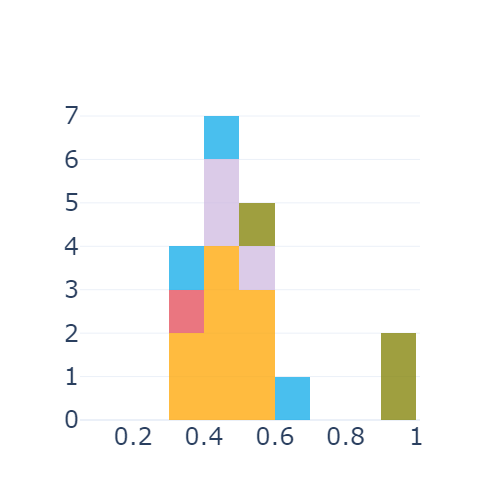}
         \caption{Hospital}
         \label{fig:specifHos}
     \end{subfigure}
     \begin{subfigure}[b]{0.31\textwidth}
     \includegraphics[width=\textwidth]{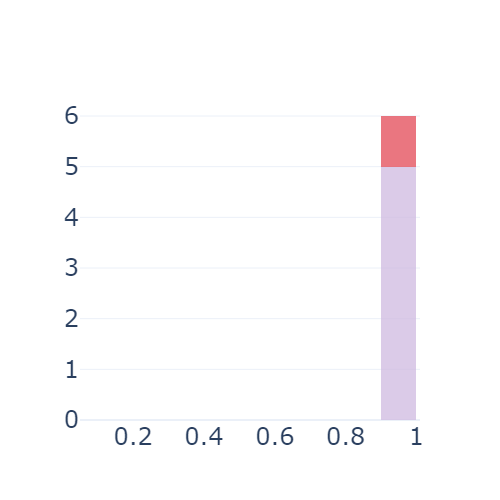}
         \caption{Primary School}
         \label{fig:specifPS}
     \end{subfigure}
     \begin{subfigure}[b]{0.35\textwidth}
     \includegraphics[width=\textwidth]{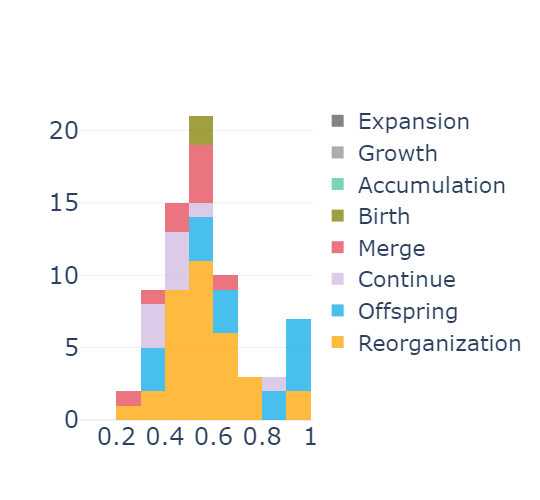}
         \caption{High School}
         \label{fig:specifHS}
     \end{subfigure}
     
        \caption{Typicality distribution of three SocioPatterns datasets using the same daily aggregation window}
     
        \label{fig:typicality}
\end{figure}

\begin{table}[ht]
\centering
\begin{tabular}{llll} 
\toprule
Duration & $\eta$ PS  & $\eta$ Hospital & $\eta$ HS \\
\midrule
1m   & 0.3592 & 0.5513 & 0.5407 \\
15m  & 0.3813 & 0.2770 & 0.4167 \\
30m  & 0.4042 & 0.1999 & 0.3276 \\
1h   & 0.4555 & 0.1579 & 0.2147 \\
2h   & 0.3674 & 0.1380 & 0.1410 \\
6h   & 0.4533 & 0.0929 & 0.0818 \\
12h  & 0.3250 & 0.1246 & 0.0471 \\
24h  & 0.7764 & 0.1444 & 0.1133 \\
\bottomrule
\end{tabular}
\captionsetup{width=10cm}
\caption{Stability scores for different durations, PS: Primary School, HS: High School. We remark that PS is highly stable at the daily aggregation, while the 2 others are more stable at the minute scale.}
\label{tab:stability}
\end{table}

\subsection{Events Stability and Typicality}
A specificity of the SocioPatterns dataset is that they are provided at a fine temporal scale of 20 seconds, but are usually studied by choosing an aggregation scale. Although several approaches exist to do so (e.g., \cite{cazabet2021data}), we can leverage our framework to select the most appropriate aggregation scale for partition evolution analysis. Intuitively, to be interpretable, the partition in a timestep should be as similar as possible to partitions in previous and following ones. We know that, by definition, the \textsc{Continue} event captures how much groups stay unchanged from one timestep to the next. 
We can thus compute a stability score $\eta$ as follows:

\[
\eta=\frac{1}{|E|}\sum_{e\in E}\textsc{Continue}(e)
\]
with $E$ the set of all events and $\textsc{Continue}(e)$ the event weight for event type $\textsc{Continue}$. Moreover, $\eta\in [0,1]$, a higher score corresponding to more stable groups. In Table \ref{tab:stability}, we observe that different SocioPatterns datasets are stable at different timescales. For instance, the Primary School dataset is the most stable of all when using a daily aggregation window. When we check the details of those stable groups, we observe that they match the primary school classes well. The daily aggregation window is not particularly stable for the two other datasets. On the contrary, it is at the smallest aggregation window, at the minute level, that groups are the most stable. When looking at the details, we observe that there are many small groups at this scale, which persist unchanged for some time steps, probably corresponding to interactions in small groups. Thanks to this result, we will use some of the most stable aggregation scales in the following.



\begin{figure}
     \centering
     \begin{subfigure}[b]{0.7\textwidth}
     \includegraphics[width=\textwidth]{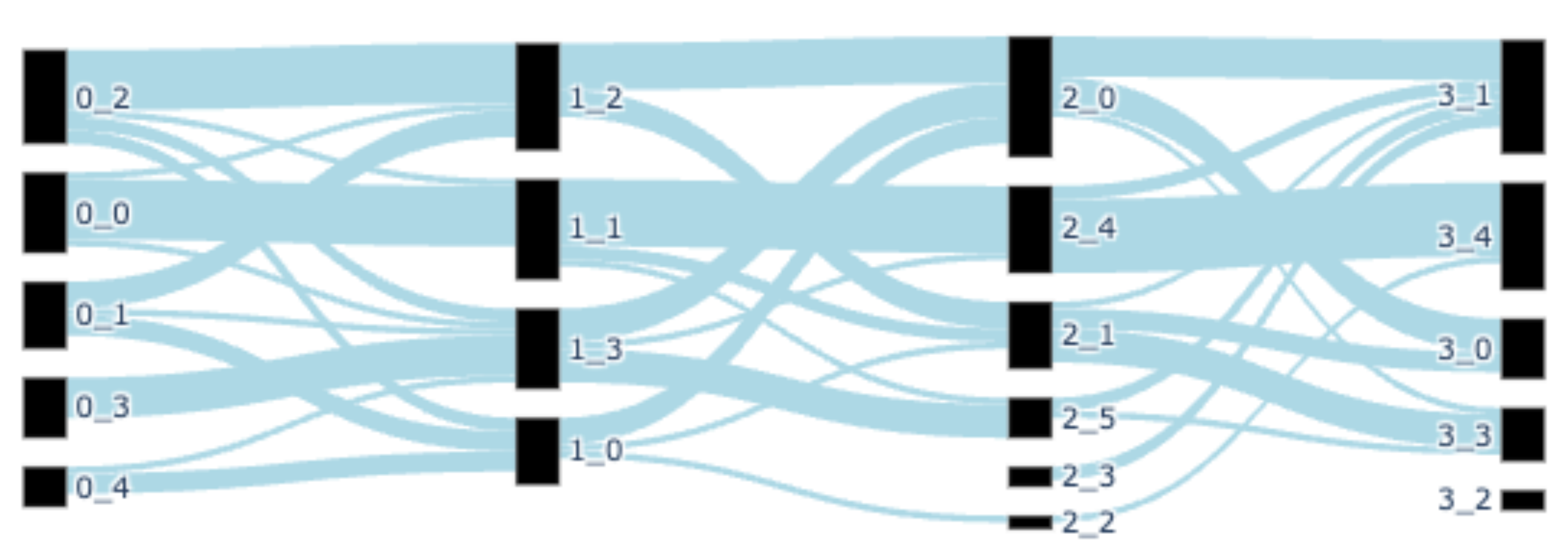}
         \caption{Entity flow between groups}
         \label{fig:hosDailyflow}
     \end{subfigure}

    \begin{subfigure}[b]{0.7\textwidth}
     \includegraphics[width=\textwidth]{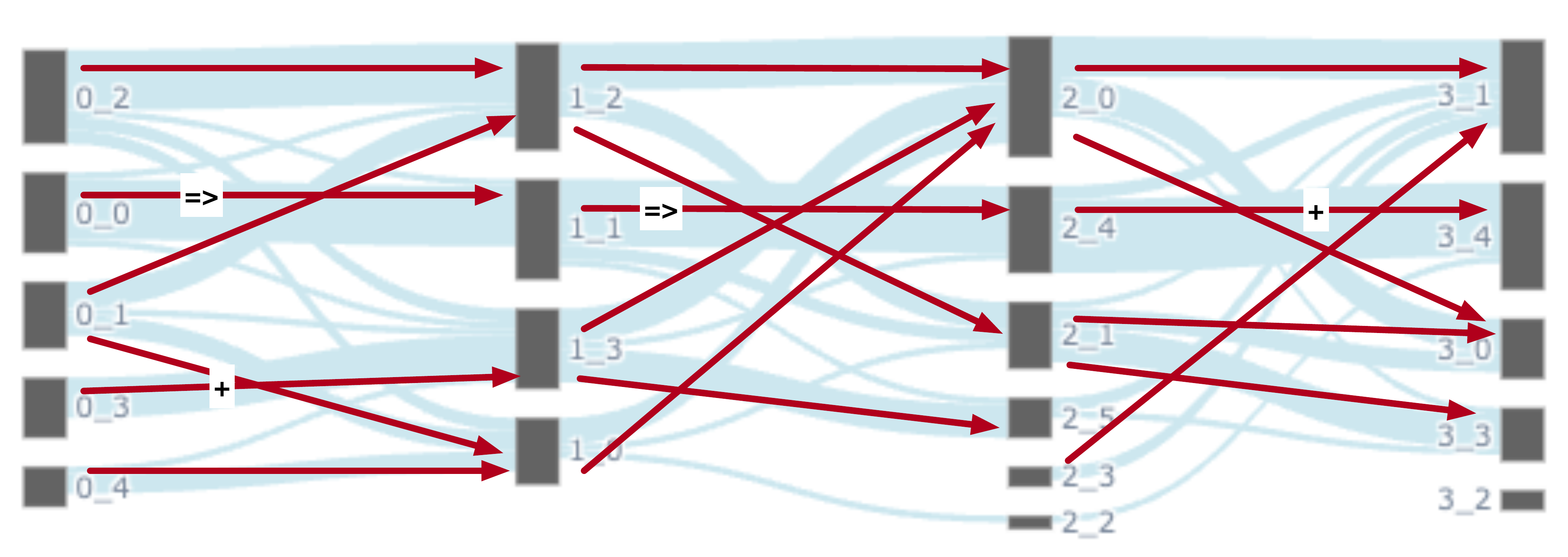}
         \caption{Event-graph according to \cite{greene2010tracking}, with $\mu=0.1$}
         \label{fig:hosDailyEG}
     \end{subfigure}

        \caption{Hospital dataset, daily windows. Groups are labeled as $t$\_$id$, where $t$ is the timestamp and $id$ is the arbitrary group id. Event legend: $=>$: \textsc{Continue}, $+$: \textsc{growth}.}
     
        \label{fig:hosDaily}
\end{figure}

In Section \ref{typicality}, we defined the typicality of an event as the maximal event score for that event in a particular direction. We can leverage this information to describe system-wide the distribution of event types and typicality in a system. In Fig. \ref{fig:typicality}, we plot this distribution for the three datasets at the daily aggregation level. The results are coherent with what we observed for the stability, with the primary school dataset composed mostly of archetype \textsc{Continue} and \textsc{Merge}, while the other datasets, mostly unstable, exhibit poorly defined events, at the exception of a few \textsc{Birth} and \textsc{Offspring}.


\subsection{Comparison with state of the art}
Comparing the quality of events detected using our approach with those found by previous methods is difficult, due to (i) the absence of external ground truth, and (ii) the difference in nature between events found by previous methods and those produced by our approach.

Therefore, we illustrate the interest of our formalism in case studies. 

\subsubsection{Hospital Dataset, Daily snapshots}

We first compare the three frameworks on the Hospital Dataset, using a daily aggregation window. Daily aggregation seems intuitively reasonable on SocioPatterns datasets, and has commonly been used in the literature. We have seen from Tab. \ref{tab:stability} that it leads to rather unstable partitions for the Hospital dataset. Fig. \ref{fig:hosDailyflow} represents the flow for the four days. We observe that the relations between groups are complex, and that assigning archetypes to events does not seem obvious.

Fig. \ref{fig:hosDailyEG} represents the event graph obtained from Greene et al.~\cite{greene2010tracking} at the $\mu=0.1$ threshold, observed to work best by the authors. We observe that in most cases, the events recognized are rather logical: \texttt{0\_2} and \texttt{2\_0} being \textsc{Merge} events, \texttt{0\_1} and \texttt{1\_3} being \textsc{Split} events, etc. Only 2 events are \textsc{Continue}, and 2 others are \textsc{Growth}. However, in several other cases, the events obtained are more disputable. Is there no relation between \texttt{1\_0} and \texttt{2\_2}? Is \texttt{1\_3} really a growth from \texttt{0\_3}? Why \texttt{0\_2} is not a split? These limits are consequences of the research of archetypes only, without taking into account the multiple facets of events. Another limit is that most events are intertwined split-and-merge, i.e., out-going branches of \textsc{Split}s form in-going branches of \textsc{Merge}s. The interpretation of such events thus becomes difficult: it seems abusive to say that, for instance, \texttt{3\_0} is a merge of \texttt{2\_0} and \texttt{2\_1}, while both these groups are subject to \textsc{Split}, and only minor fragments of them join into \texttt{3\_0}.

Using the framework from Asur et al. \cite{asur2009event} leads to an even more unsatisfactory result, since it yields only six recognized events: two \textsc{Merge}s, two \textsc{Split}s, and two \textsc{Birth}s. Indeed, this framework is very conservative in its definition of events, and most flows are too complex to be recognized as such. 

Using our framework, we obtain a richer description of events. We first can consider the typicality of events to identify the closest to archetypes. For instance, the 3 highest forward-\textsc{Continue} are \texttt{0\_0,1\_1,2\_4}, corresponding to the same group of people continuing in the same group, although not in perfect \textsc{Continue}. Similarly, the three highest \textsc{Split} scores are \texttt{1\_3,2\_1,2\_0}. We observe that these descriptions are compatible with the description obtained from Greene et al.~\cite{greene2010tracking}. However, our framework offers more details. For instance, the forward \textsc{Continue} of \texttt{0\_0} also has, in decreasing magnitude, \textsc{Ancestor}, \textsc{Split} and \textsc{Disassemble} facets. Indeed, some of its components join other groups, while its components represent only 9 out of 15 elements in \texttt{1\_1}, making it far from an archetypal \textsc{continue}.

In another example, we focus on an event that seems unconvincing according to Greene et al.~\cite{greene2010tracking}, \texttt{1\_3} seen as a growth from \texttt{0\_3}. We provide the complete facets of \texttt{1\_3-backward}:
\begin{itemize}
    \item \textsc{Birth}: 0.02
    \item \textsc{Accumulation}: 0.03
    \item \textsc{Growth}: 0.01
    \item \textsc{Expansion}: 0.02
    \item \textsc{Continue}: 0.14
    \item \textsc{Merge}: 0.25
    \item \textsc{Offspring}: 0.19
    \item \textsc{Reorganization}: 0.34
\end{itemize}
We observe that the dominant facet is \textsc{reorganization}, since it is composed of multiple minor parts of previous groups. It also has some elements of \textsc{Merge}, \textsc{Offspring}, and \textsc{Continue}, which is due to the complexity of receiving a large fraction of its components from the bulk of a single group, while the rest are minor fragments from various other groups.

From this example, we see that when the groups are subject to complex relationships, state-of-the-art approaches yield potentially misleading results, while the richer description of our framework allows a better characterization of group evolution.

\begin{figure}
     \centering

         \begin{subfigure}[b]{0.8\textwidth}
     \includegraphics[width=\textwidth]{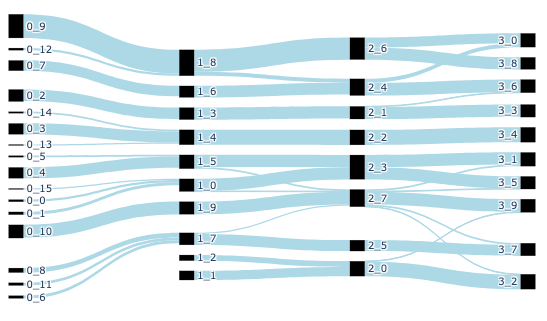}
         \label{fig:1hprimary}
     \end{subfigure}
     
        \caption{Primary School, 1h windows, first 4 hours.}
     
        \label{fig:PS1h}
\end{figure}

\begin{figure}
    \centering
     \begin{subfigure}{\textwidth}
         \centering
         \includegraphics[width=0.87\textwidth]{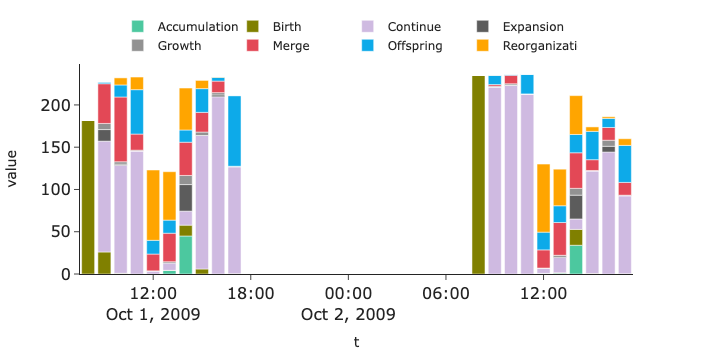}
         \caption{Backward events}
         \label{fig:sp_backward}
     \end{subfigure}
    \begin{subfigure}{\textwidth}
         \centering
                  \includegraphics[width=0.87\textwidth]{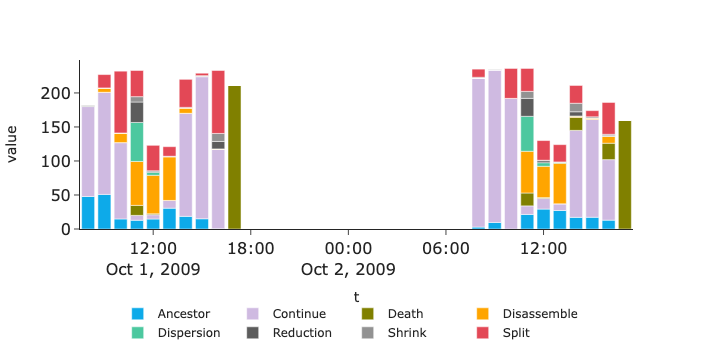}
         \caption{Forward events}

         \label{fig:sp_forward}
     \end{subfigure}
     \caption{Temporal distribution of Backward (top) and Forward (bottom) events in the Primary School dataset with a 1-hour resolution. We observe a similar pattern for the two days of analysis, with stable communities ---large fractions of \textit{continue} events--- during teaching hours, and more diverse events during lunch time}
          \label{fig:sp-time}
\end{figure}

\subsubsection{Primary School, Hourly snapshots}
We then focus on the Primary School dataset, at the hourly aggregation timescale. We have seen from Table \ref{tab:stability} that it is the most stable aggregation step, if we ignore the 1-minute aggregation steps that yield very small groups, and the PS 1-day aggregation window, which contains nearly only perfect continue events as seen in Fig. \ref{fig:typicality}. 

We first focus on the details of the flow during the first four hours and plot the details in Fig. \ref{fig:PS1h}. We observe that events seem indeed much easier to characterize than in the previous case.

 Many events are recognized as pure \textsc{Continue} by all three frameworks, such as \texttt{(0\_2$\rightarrow$1\_3)},\texttt{(0\_10$\rightarrow$1\_9)},\texttt{(1\_3$\rightarrow$2\_1)}, etc.
 All frameworks also agree on \textsc{Birth} events for \texttt{$\rightarrow$1\_1, $\rightarrow$1\_2)}, while a few other events are recognized without ambiguity as \textsc{Merge} by all frameworks, such as \texttt{(1\_2,1\_1)$\rightarrow$2\_0} or \texttt{(1\_5,1\_0)$\rightarrow$2\_3)}.


However, we also find ambiguous cases in this dataset. Let us focus on one example, the event involving groups 
\texttt{(0\_15,0\_0,0\_1,1\_0)}: According to Asur et al. \cite{asur2009event}, this event has no label. According to Greene et al., it is a merge from \texttt{0\_0,0\_1} into \texttt{1\_0}, and \texttt{0\_15} is a death. With our framework, \texttt{0\_15,0\_0,0\_1} are classified as \textsc{Ancestor}s ($\mathcal{T}=0.8$), while \texttt{1\_0}-backward is \textsc{Merge}($\mathcal{T}=0.45$) and \textsc{Expansion}($\mathcal{T}=0.45$). Indeed, we see that this new group is composed of one-half of a perfect merge of those 3 predecessors, the other half being new entities. Through this example, we see that using a richer description is also useful even when group evolution is mostly stable.

\vspace{1cm}
\noindent Beyond focusing on individual events, our framework can be used to describe system-wide dynamics.
Figure \ref{fig:sp-time} details the evolution of events for each hour, computed for each group $G$ and event score $E(G)$ for group $G$ as $|G|E(G)$.

From a backward perspective, most groups are born in the first hours (8:00-10:00), when students arrive at their classrooms and have the chance to connect.
Conversely, group death in the forward perspective occurs at the end of days, when students leave the school.

Communities are mostly stable during class hours, as exemplified by the large \textsc{Continue} areas between 9:00-12:00, and 14:00-17:00.
However, some \textsc{Merge}, respectively \textsc{Split} events can also be observed during that period. 

In the middle of the school day, students are primarily involved in \textsc{Reorganization} (backward) and \textsc{Disassemble} (forward) events, i.e., few students detach from large communities and form new groups, themselves composed of students from multiple groups. This can be interpreted as a lunchtime activity, in which students can meet peers from other classes. 
Indeed, we can validate this hypothesis by looking at the Metadata facet, using class membership as an attribute. We observe an average $\sim 25\%$ increase after \textsc{Merge} events, meaning that communities resulting from these events are more varied with respect to the contributing ones in the corresponding previous timestamp. 

We can also observe a symmetric phenomenon, with \textsc{Dispersion} events occurring in the forward direction at the beginning of the lunch break, answered with \textsc{Accumulation} (backward)  after the break. These event types are in the same color in the figures because they correspond to the same combination of facets. These events are characterized by important outflows. Indeed, one can observe that many students do not participate in the lunch break at school, which can explain this situation.

\begin{figure}[!t]
    \centering
    \includegraphics[scale=0.24]{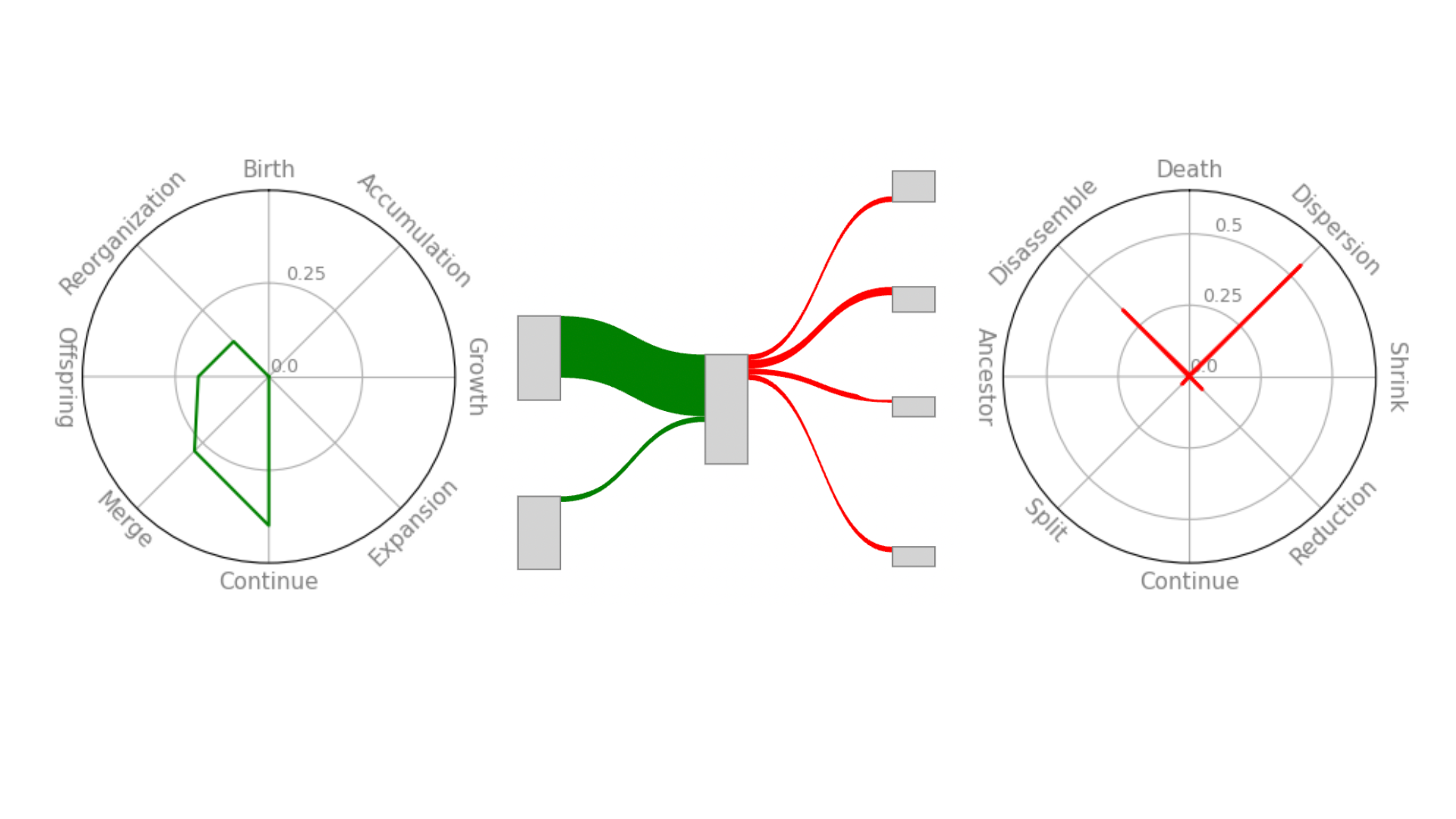}
    \caption{Backward and Forward event weights extracted from community~3\_6 in Primary School, 1-hour resolution. }
    \label{fig:radar}
\end{figure}

In fact, the full picture of a group's transformations can be obtained by analyzing all of its event weights.
For example, in Figure \ref{fig:radar}, we show backward and forward event weights for community \texttt{3\_6}.
The backward event weights highlight a \textsc{Continue}-like event showing some traits of a \textsc{Merge} (i.e., two contributing sets coming together), of an \textsc{Offspring} (due to the largest contributing community not coming in full), and of \textsc{Reorganization} (due to the low contribution of the smaller contributing community).
The forward event weights, instead, describe an event showing traits of \textsc{Disassemble}, \textsc{Dispersion}, and \textsc{Reduction} due to the disappearance of most individuals and the separation of the remaining ones into multiple, smaller, groups.
\\ \ \\

\section{Discussion and Conclusion}
\label{sec:disc_concl}
In this paper, we introduced a framework to describe the temporal dynamics of groups by assessing their similarity to archetypal categories. 
This produces a flexible frame that removes the need for arbitrary thresholds while accurately representing the transformations at play at the same time.
We have shown how this framework can be applied to the longitudinal analysis of temporal data, specifically in studying (i) the evolution of target groups, (ii) the evolution of the whole system in terms of its mesoscale dynamics, and (iii) how temporal granularity may impact observations on group transformations. We observed that, whatever the framework, identifying events with unstable dynamic groups seem doomed to fail, thus emphasizing the importance of smoothed dynamic clustering \cite{cazabet2020evaluating}.

Moreover, by introducing measures to quantify the change in the entities' labels (see Definition \ref{def:attrent}), we also suggest that the relation between group structures and entities' metadata could be explored further.
For instance, this framework could be applied to the analysis of spatiotemporal clusters in human mobility data to understand whether differences in individual attributes are related to substantial differences in group mobility patterns.
Another interesting applicative scenario could be the characterization of groups emerging from online social interactions to evaluate peer pressure effects, i.e., to measure the extent to which groups induce opinion change or, conversely, evaluate how opinion change drives group formation.
To do so, however, one should account for time-changing attribute values, an aspect we did not cover in this work.
Moving away from simply finding evolutive patterns in the data, this framework and its measures could be exploited to forecast both individual and group activity, e.g. using the proposed measures as input for a classification task.




\section*{Statements and Declarations}
\subsection*{Author Contributions}
All authors designed research, performed research, and wrote the paper. 
A.F., S.C., and R.C. contributed to the acquisition, analysis, and interpretation of the experiments.
A.F., S.C., and R.C. contributed to the software design and to its implementation. 
S.C., G.R., and R.C. coordinated and supervised all of the research.
All authors read and approved the final manuscript.

\subsection*{Funding}
This project was partly funded by: (i) BITUNAM grant ANR-18-CE23-0004; (ii) SoBigData.it which receives funding from the European Union – NextGenerationEU – National Recovery and Resilience Plan (Piano Nazionale di Ripresa e Resilienza, PNRR) – Project: “SoBigData.it – Strengthening the Italian RI for Social Mining and Big Data Analytics” – Prot. IR0000013 – Avviso n. 3264 del 28/12/2021; (iii) EU NextGenerationEU programme under the funding schemes PNRR-PE-AI FAIR (Future Artificial Intelligence Research).

\subsection*{Availability of Data and Material}
All data used in this paper is available on the SocioPatterns project website (URL: \url{www.sociopatterns.org}). No new data was generated.

\subsection*{Code Availability}
Code will be made available in a dedicated GitHub repository upon acceptance of the manuscript.

\subsection*{Competing Interests}
The authors have no relevant financial or non-financial interests to disclose.



\bibliography{sn-bibliography}

\end{document}